\documentclass[10pt,doublecolumn]{IEEEtran}
\usepackage{amsmath,amsthm,amsfonts,amssymb,bm,mathrsfs}
\usepackage{graphicx,subfig}
\usepackage{multicol}
\usepackage{mathrsfs}
\usepackage{algorithm}
\usepackage{algorithmic}
\usepackage{booktabs,multirow,array,longtable,threeparttable}
\usepackage{color}
\graphicspath{{fig/}}
\usepackage{cite,url}
\allowdisplaybreaks[4]
\begin{document}

\title{Deep Reinforcement Learning with Discrete Normalized Advantage Functions for Resource Management in Network Slicing}

\author{
Chen Qi, Yuxiu Hua, Rongpeng Li, Zhifeng Zhao, and Honggang Zhang

\thanks{The authors are with Zhejiang University, Zheda Rd. 38, Hangzhou 310027, China (mail: \{qichen7c, 21631087, lirongpeng, zhaozf, honggangzhang
\}@zju.edu.cn). Corresponding author: R. Li.}

\thanks{This work was supported by National Natural Science Foundation of China (No. 61701439, 61731002), Zhejiang Key R\&D Plan (No. 2019C01002), the Fundamental Research Funds for the Central Universities.}
}

\maketitle

\begin{abstract}
Network slicing promises to provision diversified services with distinct requirements in one infrastructure. Deep reinforcement learning (e.g., deep $\mathcal{Q}$-learning, DQL) is assumed to be an appropriate algorithm to solve the demand-aware inter-slice resource management issue in network slicing by regarding the varying demands and the allocated bandwidth as the environment \textit{state} and the \textit{action}, respectively. However, allocating bandwidth in a finer resolution usually implies larger action space, and unfortunately DQL fails to quickly converge in this case. In this paper, we introduce discrete normalized advantage functions (DNAF) into DQL, by separating the $\mathcal{Q}$-value function as a state-value function term and an advantage term and exploiting a deterministic policy gradient descent (DPGD) algorithm to avoid the unnecessary calculation of $\mathcal{Q}$-value for every state-action pair. Furthermore, as DPGD only works in continuous action space, we embed a k-nearest neighbor algorithm into DQL to quickly find a valid action in the discrete space nearest to the DPGD output. Finally, we verify the faster convergence of the DNAF-based DQL through extensive simulations.
\end{abstract}

\IEEEpeerreviewmaketitle

\section{Introduction}
Networks are becoming increasingly agile and flexible to provision diversified services with distinct requirements on latency and rate. Specifically, network slicing, which belongs to one of cutting-edge technologies in the 5G era, allows infrastructure providers to offer ``slices" of resources (computational, storage and networking) with specified service license agreements (SLAs) \cite{zhou_network_2016, li_intelligent_2017, li_deep_2018,zheng_elastic_2019}. However, in order to fully reap the desired merits like slice-level protection, envyfreeness, and load-driven elasticity \cite{li_deep_2018,zheng_elastic_2019}, end-to-end network slicing still faces a lot of technical challenges. For example, taking account of the limited spectrum, the slice-level protection could guarantee superior quality of experience (QoE) but also incurs degradation in spectrum efficiency (SE). Therefore, one typical question naturally arises like that how to intelligently allocate the spectrum to slices according to the dynamics of service request from mobile users in a coherent manner\cite{vassilaras_algorithmic_2017}, so as to obtain satisfactory QoE in each slice at the cost of acceptable SE.

In order to address the aforementioned problem, one potential solution is to consider the (deep) reinforcement learning (RL). As a non-nascent algorithm, RL has been widely applied in the field of cognitive radio \cite{chen_stochastic_2013} and green communications \cite{li_tact:_2014}. The recent well-known application success in Go \cite{mnih_human-level_2015} further proves the feasibility to utilize neural networks (NN) to approximate the value functions in classical RL with case-testified convergence stability, and triggers tremendous research attention in the communications and networking area to solve resource allocation issues in some specific fields like power control, green communications, cloud radio access networks, mobile edge computing and caching \cite{li_deep_2018}. But, a common problem in these works is that researchers usually assume a rather limited small discrete action space to ensure the necessary convergence rate. For example, \cite{li_deep_2018} realized the spectrum allocation per slice on the unit of MegaHertz and accordingly design a DRL framework with tens of possible actions. But, such a coarse-grained spectrum allocation solution inevitably decreases the SE when some slice has very few service activities. In a word, it urgently needs a rethink on DRL to better avoid the curse of dimensionality and quickly converge in larger action space.

Overall speaking, this paper aims to answer how to allocate the limited spectrum on a finer-grained resolution across slices based on an improved DRL. In particular, we revolutionize the calculation and approximation of the $\mathcal{Q}$-value function in the deep $\mathcal{Q}$-learning (DQL) as follows:
\begin{itemize}
\item Inspired by the normalized advantage functions (NAF) model \cite{gu_continuous_2016,wang_dueling_2015}, we design a discrete NAF (DNAF) NN to separately approximate a state-value function term $V(\bm{s})$ and an advantage term $A(\bm{s}, \bm{a})$, where $\bm{s}$ and $\bm{a}$ denote a state and an action respectively. Moreover, we have $Q(\bm{s}, \bm{a}) = V(\bm{s}) + A(\bm{s}, \bm{a})$. Hence, the common part of the $\mathcal{Q}$-value function could be learnt across all actions.
\item We utilize a deterministic policy gradient descent (DPGD)-based $\mathcal{Q}$-learning \cite{silver_deterministic_2014} to replace the classical statistical policy gradient descent-based DQL, so as to directly yield the most suitable action for a specific state. 
\item In order to solve the issue that a DPGD method ignores the discreteness of the action space, we firstly output a proto action with the largest $\mathcal{Q}$-value in the virtual continuous action space and then scramble it with an extra noise term. Finally, we embed a k-nearest neighbor (k-nn) algorithm to quickly select the closest discrete action.
\end{itemize}
The remainder of the paper is organized as follows: Section \ref{sec:model} talks about some necessary algorithmic background and formulates the system model. Section \ref{sec:algorithm} gives the details of the DNAF based $\mathcal{Q}$-Learning, while Section \ref{sec:results} presents the related simulation results. Finally, Section \ref{sec:conclusion} concludes the paper with a summary.

\section{Mathematical Background and System Model}
\label{sec:model}
\subsection{Mathematical Background}
RL tries to find a strategy $\pi$, which maps a state $\bm{s} \in \mathbb{S}$ (i.e., the varying traffic per slice) to an action $\pi(\bm{s})$, (i.e., $\bm{a} \in \mathbb{A}$, allocated bandwidth per slice) to maximize the discounted accumulative reward starting from the state $\bm{s}^{(0)}= \bm{s}$. Formally, this accumulative reward is called as a state-value function, which can be calculated by \cite{sutton_reinforcement_1998}
\begin{equation}
\label{eq:definition_state_value_function}
V^{\pi}(\bm{s}) =\mathbb{E}_{\pi}\left[\sum\limits_{k=0}^{\infty} \gamma^k R(\bm{s},\pi(\bm{s})|\bm{s}^{(0)}=\bm{s})\right]
\end{equation}
where the positive parameter $\gamma$ is the discount factor that maps the future reward to the current state. Given the diminishing importance of future cost than the current one, $\gamma$ is smaller than 1.

$\mathcal{Q}$-learning is an RL technique to obtain the strategy $\pi$. Specifically, a $\mathcal{Q}$-learning agent attempts to learn the value of taking a specific action for a given state, i.e. $Q$-value, by constantly updating $Q$ value in a temporal difference (TD) manner as
\begin{align}
	& Q(\bm{s}, \bm{a}) \\
	\gets & Q(\bm{s}, \bm{a}) + \alpha \big(R(\bm{s}, \bm{a}) + \gamma\max_{\bm{a'} \in \mathbb{A} }Q(\bm{s'}, \bm{a'}) - Q(\bm{s}, \bm{a}) \big) \nonumber
\end{align}
where $\alpha$ is the learning rate and $\bm{s'}$ denotes the state of the environment after taking action $\bm{a}$ at state $\bm{s}$. 


In recent years, \cite{mnih_human-level_2015} proposes to use NNs to approximate $\mathcal{Q}$-value function \cite{hornik_multilayer_1989} so as to solve an RL problem with a tremendous state dimension. Mathematically, DQL trains an NN with parameters $\bm{\theta}$ by minimizing the loss function between the real $\mathcal{Q}$-value function $Q(\bm{s},\bm{a})$ and an NN-approximated one $Q^{+}(\bm{s},\bm{a}|\bm{\theta})$, which can be formulated as
\begin{align}
\bm{\theta} = \arg \min_{\bm{\theta'}}	L(\bm{\theta'}) = \arg \min_{\bm{\theta'}} \big(Q(\bm{s},\bm{a}) - Q^{+}(\bm{s},\bm{a}|\bm{\theta'}) \big)^2
\end{align}
Commonly, $\bm{\theta}$ could be achieved by a gradient-based approach as
\begin{align}
	\label{eq:gradient}
	\bm{\theta}
	\leftarrow  \bm{\theta} - \alpha \nabla L(\bm{\theta}) 
\end{align}
In addition, there are some tricks that can improve the performance of DQL, such as replay buffer \cite{lillicrap_continuous_2015}, target network \cite{hasselt_deep_2016}, prioritized replay \cite{schaul_prioritized_2015}, etc. 

\subsection{System Model}
\begin{figure*}
	\centering
	\includegraphics[width=.675\textwidth]{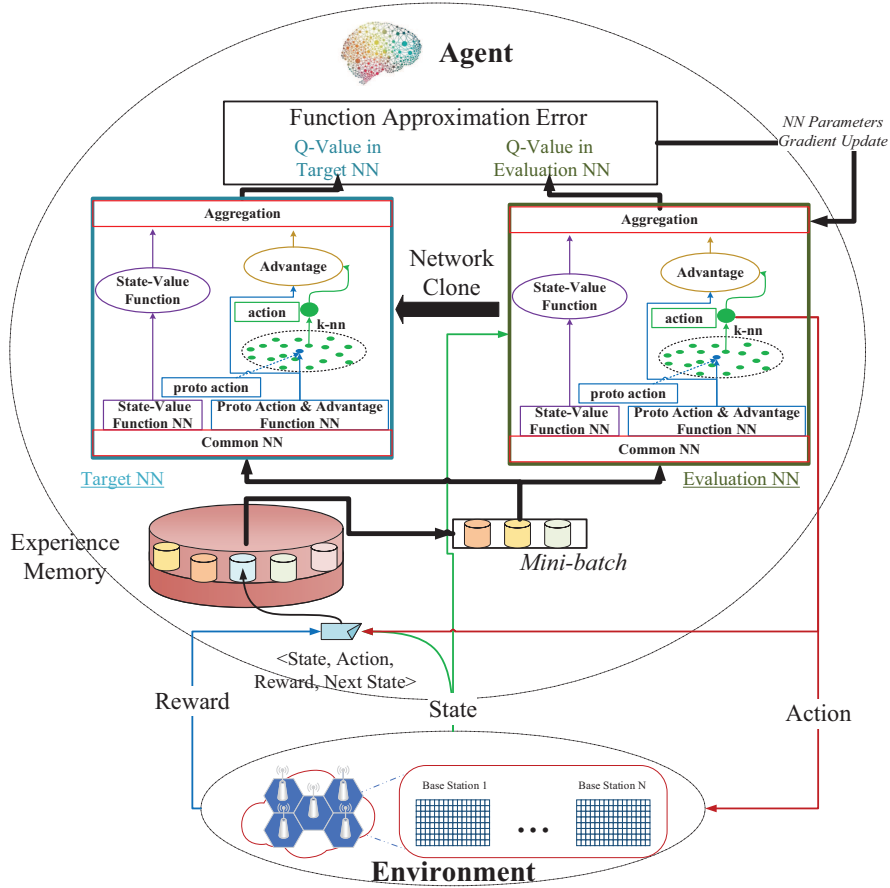}
	\caption{An illustration of DNAF-based $\mathcal{Q}$-learning for resource management in network slicing.}
	\label{fig:DNAF} 
\end{figure*}

We consider an access network scenario in Fig. \ref{fig:DNAF} consisting of multiple base stations (BSs) , where there exists a list of existing slices $1, \cdots, N$ sharing the aggregated bandwidth $W$ and having fluctuating demands $\bm{d}=(d_1, \cdots, d_N)$. We aim to maximize the expectation of the utility function $\mathbb{E}\{R(\bm{w},\bm{d})\}$, where the notation $\mathbb{E}(\cdot)$ denotes to take the expectation of the argument. Moreover, the utility function is defined as the weighted sum of SE and QoE satisfaction ratio. Mathematically, 
\begin{equation}
	\label{eq:reward_defintion}
	R = \zeta \cdot \text{SE} + \beta \cdot \text{QoE}
\end{equation}
where $\zeta$ and $\beta$ denotes the importance of SE and QoE. Our goal is to allocate the bandwidth to slices according to the traffic variations within each slice, that is,
\begin{align}
	&\arg \max_{\bm{w}} \mathbb{E}_t \{R(\bm{w},\bm{d})\} \nonumber\\
	= & \arg \max_{\bm{w}} \mathbb{E}\big\{\zeta \cdot \text{SE} (\bm{w},\bm{d})+ \beta \cdot \text{QoE} (\bm{w},\bm{d}) \big\} \nonumber\\
	\text{s.t.: } & \bm{w}=(w_1, \cdots, w_N) \label{eq:formulation}\\
	& w_1+ \cdots + w_N = W \nonumber\\	
	& w_i = k \cdot \Delta, \forall i \in [1, \cdots, N] \nonumber\\	
	& \bm{d}=(d_1, \cdots, d_N )\nonumber\\
	& d_i \sim  \text{Certain Traffic Model}, \forall i \in [1, \cdots, N] \nonumber
\end{align}
where $t$ denotes the temporal index, $k$ is an integer and $\Delta$ is the minimum allocated bandwidth per slice. Notably, $\bm{d}(t)$ depends on both $\bm{d}(t-1)$ and $\bm{w}(t-1)$, since the maximum sending capacity of servers belonging to one service is tangled with the provisioning capabilities for this service. For example, the TCP sending window size is influenced by the estimated channel throughput.

The key challenge to solve \eqref{eq:formulation} lies in the volatile demand variations without having known a priori due to the traffic model. Hence, DQL promises to be an appropriate solution to solve this problem. But, DQL converges slowly for large action space, since DQL needs to predict $\mathcal{Q}$-values for each state-action pair. Unfortunately, the size of action space $\vert \mathbb{A} \vert$ increases exponentially along with the decrease of $\Delta$ and the increase of $N$, since we have 
\begin{align}
	\vert \mathbb{A} \vert = \binom{\lfloor\frac{W}{\Delta} \rfloor -1 }{N-1}
\end{align}
Therefore, it is inevitable to revolutionize the classical DQL.

\section{The DNAF based $\mathcal{Q}$-Learning}
\label{sec:algorithm}
 Researchers in \cite{wang_dueling_2015,gu_continuous_2016} has suggested NAF as a potential solution to DQL with continuous action space by decomposing $\mathcal{Q}$-value into a state-value function $V$ and an advantage function $A$. Since the discreteness of action space for resource management in network slicing is different from the continuity in \cite{wang_dueling_2015,gu_continuous_2016}, it is quite meaningful to re-investigate its effectiveness here.

Inspired by \cite{wang_dueling_2015,gu_continuous_2016}, we build two separate NNs for the state-value function $V(\bm{s}|\bm{\theta}^V) $ parameterized by $\bm{\theta}^V$ and the advantage function $A(\bm{s}, \bm{a}|\bm{\theta}^A)$ parameterized by $\bm{\theta}^A$, on top of a common NN collecting some general information, that is,
\begin{align}
	\label{eq:q_decomposition}
	Q(\bm{s}, \bm{a}| \bm{\theta}^Q) &=  V(\bm{s}|\bm{\theta}^V) + A(\bm{s}, \bm{a}|\bm{\theta}^A)
\end{align}
Moreover, the advantage function approximation leverages the DPGD algorithm \cite{silver_deterministic_2014} to obtain the proto action $\bm{\mu}(\bm{s}|\bm{\theta}^{\bm{\mu}})$ by an parameterized by $\bm{\theta}^{\bm{\mu}}$ by 
\begin{align}
	\label{eq:advantage}
	&A(\bm{s}, \bm{a}|\bm{\theta}^A)\\
	=  & -\frac{1}{2}(\bm{a} - \bm{\mu}(\bm{s}|\bm{\theta}^\mu))^{T}\Lambda(\bm{s}|\bm{\theta}^L)\Lambda(\bm{s}|\bm{\theta}^L)^{T}(\bm{a} - \bm{\mu}(\bm{s}|\bm{\theta}^\mu)) \nonumber
\end{align}
where $\Lambda(\bm{s}|\bm{\theta}^L)$ is a low-triangular matrix whose entries come from a linear output layer of an NN parameterized by $\bm{\theta}^L$ \cite{gu_continuous_2016}. Hence, $\bm{\theta}^A$ is a concatenation of $\bm{\theta}^{\bm{\mu}}$ and $\bm{\theta}^L$.

Since the action $\bm{\mu}(\bm{s}|\bm{\theta}^{\bm{\mu}})$ might be an invalid action in the discrete action space $\mathbb{A}$, we first calculate the proto action $\hat{\bm{a} } $\footnote{As depicted in Fig. \ref{fig:DNAF}, the proto action is plotted in the blue circle while the valid actions are presented in green ones.} by adding $\bm{\mu}(\bm{s}|\bm{\theta}^{\bm{\mu}})$ with a noise term. Notably, the added noise in selecting action procedure could be regarded as an alternative means to the $\epsilon$-greedy strategy in classical RL for state exploration. Afterwards, k-nn is applied to find the closest valid action, that is,
\begin{align}
\label{eq:knn}
g_k(\bm{a}) = \arg \min\nolimits_{\bm{a}'\in\mathbb{A}} \left\| \bm{a}' -  \hat{\bm{a} } \right\|_2
\end{align}
In other words, the function $g_k ( \cdot )$ is a k-nn mapping from a continuous space to a discrete set and returns $k$ valid actions closest to the proto action. For $k > 1$, it also means to obtain the $k$ nearest actions that maximize $\mathcal{Q}$-value\cite{dulac-arnold_deep_2015}. The case $k = 1$, which belongs to the key focus in this paper, is equivalent to simply select the nearest action. 

We incorporate the aforementioned methods into the DQL and have the DNAF-based $\mathcal{Q}$-learning  in Algorithm \ref{al:alg1}.

\begin{algorithm}
	\caption{The DNAF based $\mathcal{Q}$-learning} 
	\label{al:alg1}
	\begin{algorithmic}[1]
		\STATE Randomly initialize a normalized $\mathcal{Q}$ network $Q(\bm{s}, \bm{a}|\bm{\theta}^Q)$, where $\bm{\theta}^Q$ is a concatenation of $\bm{\theta}^V$ and $\bm{\theta}^A$.
		\STATE Initialize a target network $\mathcal{Q}'$ with weight ${\bm{\theta}^{Q'}}\gets{\bm{\theta}^Q}$.
		\STATE Initialize replay buffer $\mathbb{R} \gets\varnothing$.
		\STATE Initialize an episode index $t=0$.
		\REPEAT
		  \STATE At episode $t$, the agent observes the state $\bm{s}_t$.
		  \STATE The agent calculates a
		  proto-action $\hat{\bm{a} }_t = \bm{\mu}(\bm{s}_t|\bm{\theta}^\mu) + \mathcal{N}_t$
		  and determines the closest action $\bm{a}_t = g_1(\hat{\bm{a}}_t)$ by \eqref{eq:knn}.
		  \STATE The agent receives the reward $R(\bm{s}_t,\bm{a}_t)$ and observes a new state $\bm{s}_{t+1}$ for the environment.
		  \STATE The agent stores the episode experience $(\bm{s}_t, \bm{a}_t, R_t, \bm{s}_{t+1})$ in $\mathbb{R} $.
		  \STATE The agent samples a mini-batch  $\mathbb{R}_{\rm{mbatch}}$ of experiences from $\mathbb{R}$.
		  \STATE The agent sets $y_i = R_i + \gamma{V'(\bm{s}_{i+1}|\bm{\theta}^{Q'})}$ and gets estimated $\mathcal{Q}$-value $Q_i = Q(\bm{s}_i, \bm{a}_i|\bm{\theta}^Q))$ by (5) and (6), $\forall (\bm{s}_i, \bm{a}_i, R_i, \bm{s}_{i+1}) \in \mathbb{R}_{\rm{mbatch}}$.
		  \STATE The agent updates the weights $\bm{\theta}^Q$ for the evaluation network by leveraging gradient descent algorithm to $\frac{1}{|\mathbb{R}_{\rm{mbatch}}| }\sum_i((y_i - Q_i)^2)$.
		  \STATE The agent clones the $\mathcal{Q}$ network to the target network $\mathcal{Q}'$ every $C$ episodes by assigning the weights $\mathcal{Q}'$ as $\bm{\theta}^{Q'} = \bm{\theta}^Q$.
		  \STATE The episode index is updated by $t \leftarrow t+1$.
		\UNTIL{A predefined stopping condition (e.g., the gap between $y_i$ and $Q_i$, the episode length, etc) is satisfied.}
	\end{algorithmic}
\end{algorithm}
\section{Simulation Results and Numerical Analyses}
\label{sec:results}
In this part, we compare the convergence rate of the DNAF-based DQL and classical DQL. We simulate in one single BS scenario with three types of services (i.e., VoLTE, video, ultra-reliable low-latency communications (URLLC)) as in \cite{li_deep_2018,ngmn_ngmn_2007} and correspondingly have three slices. Moreover, we attempt to allocate $10$-MegaHertz bandwidth to these three slices and set the minimal bandwidth allocation resolution is $0.2$-MegaHertz, thus leading to 1176 valid actions. Meanwhile, the network slice stops sending new packets to one user if 5 packets in the caching buffer for this user have not been successfully delivered or expired (e.g., exceeding the tolerant delay for that slice). Otherwise, each network slice sends traffic to its user following the settings in Table \ref{tab:traffic}.

\begin{table}  
	\centering
	\caption{A Brief Summary of Key Settings for Traffic Generation Per Slice} 
	\label{tab:traffic} 
	\begin{tabular}{m{2.75cm} | m{1.5cm} | m{1.5cm}  | m{1.5cm}  } 
		\toprule[0.8pt] 
		& VoLTE & Video & URLLC  \\  
		\midrule[0.8pt] 
		Bandwidth & \multicolumn{3}{l}{10 MHz}\\
		\hline
		Scheduling & \multicolumn{3}{l}{Round robin per slot (0.5 ms)}\\
		\hline
		Slice Band Adjustment (Q-Value Update) & \multicolumn{3}{l}{1 second (2000 scheduling slots) }\\
		\hline
		Channel & \multicolumn{3}{l}{Rayleigh fading}\\
		\hline 
		User No. (100 in all) & 46 & 46 & 8\\
		\hline
		Distribution of Inter-Arrival Time per User& Uniform [Min = 0, Max = 160ms] & Truncated Pareto [Exponential Para = 1.2, Mean = 6 ms, Max = 12.5 ms] & Exponential [Mean = 180 ms] \\
		\hline
		Distribution of Packet Size & Constant (40 Byte) &  Truncated Pareto [Exponential Para = 1.2, Mean = 100 Byte, Max = 250 Byte]  & Truncated Lognormal
		[Mean = 2 MB, Standard Deviation = 0.722 MB, Maximum =5 MB] \\
		\hline 
		SLA: Rate & 51 kbps & 5 Mbps & 10 Mbps \\
		\hline
		SLA: Latency & 10 ms & 10 ms & 5 ms\\
		\bottomrule[1.5pt] 
	\end{tabular}  
	\end{table} 

\begin{figure}
	\centering
	\includegraphics[width=.475\textwidth]{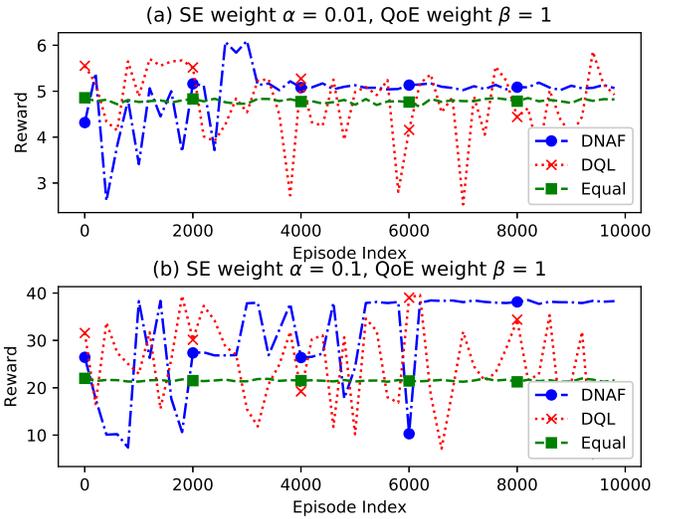}
	\caption{The convergence rate of the DNAF-based DQL and classical DQL.}
	\label{fig:reward} 
\end{figure}

\begin{figure}
	\centering
	\includegraphics[width=.475\textwidth]{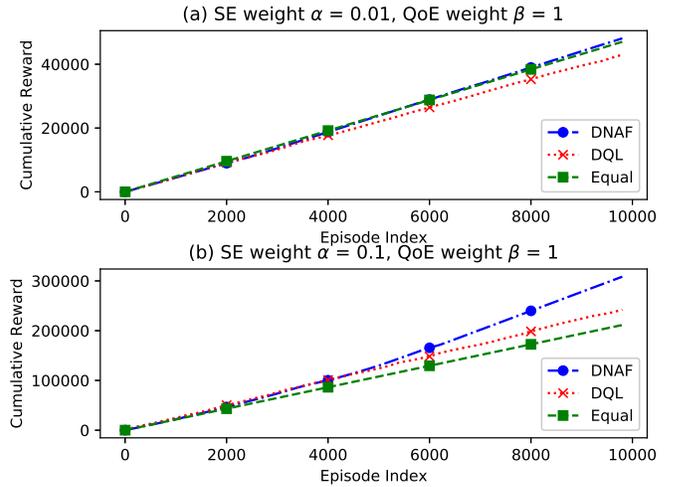}
	\caption{The cumulative reward of the DNAF-based DQL, classical DQL, equal-allocation strategy.}
	\label{fig:cumsum_reward} 
\end{figure}

Fig. \ref{fig:reward} gives an illustration of the reward variations with respect to the episode index. Here, the reward is defined in \eqref{eq:reward_defintion}. As for the DNAF-based DQL, inspired by the noise settings in \cite{lillicrap_continuous_2015}, we first assume the noise obeys normal distribution and is multiplied by an attenuating coefficient, which gradually decays with the number of iterations and ultimately fixes at zero after 3000 iterations. It can be observed from Fig. \ref{fig:reward} that regardless of the values of $\alpha$, the DNAF-based DQL could converge after $4000$ - $6000$ episodes, while the classical DQL still changes dramatically with no sign of convergence even after $10000$ episodes. Therefore, it could safely come to the conclusion that the DNAF-based DQL could converge more rapidly than the classical DQL. On the other hand, Fig. \ref{fig:reward} also provides the performance result of the equal-allocation strategy where we intuitively allocate the bandwidth according to the number of slices and verified that the DNAF-based DQL yield superior performance than the equal-allocation strategy. Furthermore, Fig. \ref{fig:cumsum_reward} gives the cumulative reward of the DNAF-based DQL, classical DQL, equal-allocation strategy. It can be found in Fig. \ref{fig:cumsum_reward} that the DNAF-based DQL could obtain superior performance than the equal-allocation strategy while DQL yield some inferior performance than the equal-allocation strategy due to its slow convergence rate.

\begin{figure}[t]
	\centering
	\includegraphics[width=.475\textwidth]{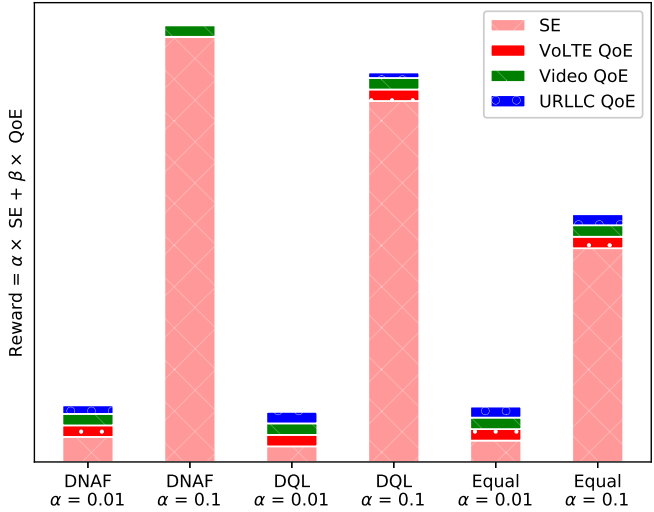}
	\caption{The snapshot of SE and QoE satisfaction ratio applying the learnt policy after 10000 episodes.}
	\label{fig:detail} 
\end{figure}

Fig. \ref{fig:detail} presents the SE and QoE satisfaction ratio applying the learnt policy after $10000$ episodes. It can be observed that when $\alpha = 0.01$, which implies that SE is on a par with QoE satisfaction ratio, the DNAF-based DQL could yield superior performance on SE and QoE satisfaction ratio simultaneously than the classical DQL. On the other hand, when $\alpha$ takes a larger value (i.e., $0.1$) to put more focus on SE, compared than the classical DQL, the DNAF-based DQL learns a policy giving significantly higher SE but degrades the QoE satisfaction ratio for some slices. Notably, some evaluation metrics produced by the DNAF-based DQL policy are not always superior (or even inferior) to the classical DQL, since it is still a very challenging research topic to design RL with multiple conflicting rewarding metrics. In this paper, we simply choose the weighted sum of two conflicting metrics (i.e., SE and QoE) as the reward in RL. Despite the intuitiveness of this direct summation, our simulation results have demonstrated that we cannot guarantee to simultaneously obtain superior performance for both SE and QoE. Therefore, we have left this inspiring and interesting topic as our future works.

\begin{figure}
	\centering
	\includegraphics[width=.475\textwidth]{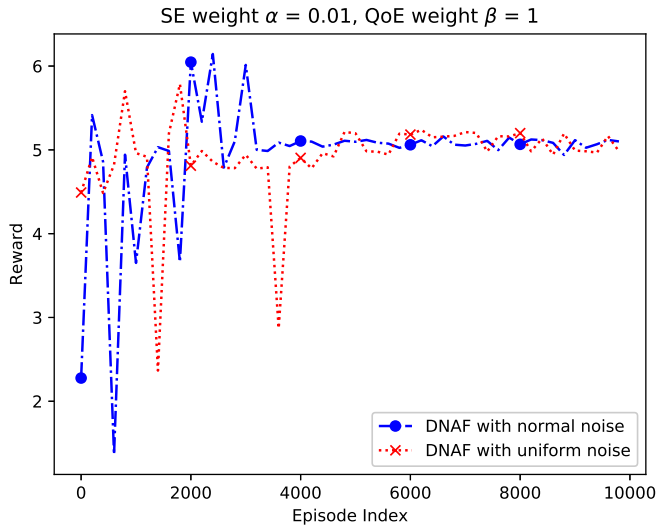}
	\caption{The reward of the DNAF-based DQL with the action noise obeying normal distribution and uniform distribution.}
	\label{fig:reward_noise} 
\end{figure}

Fig. \ref{fig:reward_noise} shows the comparison between normal distributed noise and uniform distributed ones, respectively. It can be observed that both cases could lead to convergent learning policy and exhibit trivial performance difference.

\section{Conclusion}
\label{sec:conclusion}
In this paper, we have discussed how to accelerate the convergence rate of the classical DQL in large action space, so as to satisfy the requirements for finer-resolution resource management in network slicing. In particular, we have applied the DNAF into DQL, by separating the $\mathcal{Q}$-value function as a state-value function term and an advantage term and exploiting a DPGD algorithm to avoid the unnecessary calculation of $\mathcal{Q}$-value for every state-action pair. Furthermore, we have embedded a k-nn algorithm into DQL to quickly find a valid action in the discrete action space. We have also verified that compared than the classical DQL, the DNAF-based DQL exhibits faster convergence and superior performance. Hence, we believe our works could contribute to enhancing the applicability of DQL in network slicing. However, there still exist some research issues to be solved, in particular the need to further improve DQL to guarantee minimal slice SLAs, capably adapt non-stationary traffic demands, and smartly design the reward for multi-conflicting metrics.



\end{document}